\documentclass{article}

\usepackage{microtype}
\usepackage{graphicx}
\usepackage{subcaption}
\usepackage{booktabs}
\usepackage{hyperref}
\usepackage{amsmath}
\usepackage{amssymb}
\usepackage{mathtools}
\usepackage{amsthm}
\usepackage[capitalize,noabbrev]{cleveref}

\usepackage[accepted]{icml2026}

\icmltitlerunning{Neural Estimation of Pairwise MI in Masked Discrete Sequence Models}

\begin{document}

\twocolumn[
\icmltitle{Neural Estimation of Pairwise Mutual Information in Masked Discrete Sequence Models}

\icmlsetsymbol{equal}{*}
\begin{icmlauthorlist}
\icmlauthor{Yifan Wang}{equal,berkeley}
\icmlauthor{Jai Sharma}{equal,berkeley}
\icmlauthor{Bryan Li}{equal,berkeley}
\end{icmlauthorlist}

\icmlaffiliation{berkeley}{University of California, Berkeley, CA, USA}

\icmlcorrespondingauthor{Jai Sharma}{jais@berkeley.edu}

\icmlkeywords{Mutual Information, Discrete Diffusion, Masked Generative Models, Parallel Decoding}

\vskip 0.3in
]

\printAffiliationsAndNotice{\icmlEqualContribution}

\begin{abstract}
Understanding dependencies between variables is critical for interpretability and efficient generation in masked diffusion models (MDMs), yet these models primarily expose marginal conditional distributions and do not explicitly represent inter-variable dependence. We propose a neural framework for estimating pairwise conditional mutual information (MI) directly from the hidden states of a pretrained MDM, using ground-truth MI computed from the model’s own conditional distributions for supervision. The resulting estimator captures the model’s internal belief about dependency structure and predicts the full MI matrix in a single forward pass, enabling MI-guided parallel decoding by identifying conditionally independent subsets of variables. We evaluate our approach on Sudoku and protein sequence generation with ESM-C, where the MI maps recover known structural constraints and enable a 3-5x magnitude reduction in inference-time forward passes compared to sequential decoding, while preserving generative quality and outperforming entropy-based parallelization methods.
\end{abstract}

\section{Introduction}
Understanding dependencies between elements of a discrete sequence is crucial for both interpretability and efficiency in generative models. Traditional estimation of mutual information requires density estimation, which becomes computationally intractable in high-dimensional sequences. Recent work on neural estimators of MI \cite{belghazi2018mine} motivates a framework where \textbf{pairwise MI or conditional entropy} is predicted directly from model hidden states.

In this work, we apply this framework to two domains:
\begin{enumerate}
    \item \textbf{Sudoku:} analyzing inter-cell dependencies and using MI maps to guide solving strategies.
    \item \textbf{Protein sequences (ESM-C):} interpreting position-wise dependencies and using MI to guide parallelized conditional sampling.
\end{enumerate}

\section{Literature Review}
\subsection{Discrete Diffusion and Masked Generative Models} Generative modeling for discrete data such as text has traditionally been dominated by autoregressive (AR) approaches \cite{vaswani2017attention, brown2020language}. While effective, AR models suffer from a fixed regression ordering, limiting the efficiency for long chain sequence generation such as proteins. 
\\
\\
Masked diffusion models (MDMs) are strong alternatives to AR models \cite{mdm}. Sohl-Dickstein et al. \cite{sohl2015deep} and Ho et al. \cite{ho2020denoising} established the diffusion framework for continuous data, which was later extended to discrete state spaces by Austin et al. \cite{austin2021structured} and Hoogeboom et al. \cite{hoogeboom2021argmax}. In the continuous case, noise is generated using a Gaussian distribution, while in the discrete case, noise is modeled as a transition of a token to an absorbing masked token state. 
\\
\\
 Recent works, such as Mask-Predict \cite{ghazvininejad2019mask}, demonstrate that high-fidelity sequences can be generated in constant or logarithmic steps by iteratively unmasking tokens based on model confidence. However, these methods typically rely on marginal confidence (entropy) rather than pairwise dependencies (mutual information) to determine decoding order, potentially breaking global consistency in highly structured data like Sudoku or protein folding.

\subsection{Neural Estimation of Mutual Information}
Mutual information (MI) is a fundamental quantification of the dependency between two variables, but its exact computation is often intractable in high dimensional settings. 
\\
\\
Recent advances in neural estimation of MI utilize deep learning to approximate lower or upper bounds of MI. Belghazi et al. \cite{belghazi2018mine} introduced MINE (Mutual Information Neural Estimation), which uses a neural network to maximize a lower bound on the KL-divergence. Following this, Poole et al. \cite{poole2019variational} and Cheng et al. \cite{cheng2020club} proposed variational bounds such as INFO-NCE and CLUB that offer lower variance and better stability for representation learning.
\\
\\
In the context of sequence modeling, MI is typically used for feature selection or contrastive learning such as in SimCLR \cite{chen2020simple}. Our work differs by applying MI estimation dynamically during the inference process of a generative model, predicting pairwise MI between \textit{any two indices} $i$ and $j$ given the current unmasked context $C$: $I(X_i; X_j|C)$. Through this prediction task, we explicitly model the dependency structure of the generation process.

\subsection{Adaptive and Order Agnostic Sampling}
The standard diffusion sampling process removes noise uniformly or according to a fixed schedule. However, for discrete data, joint dependencies vary across the sequence. Order-agnostic autoregressive models like XLNet \cite{yang2019xlnet} attempted to learn arbitrary factorization orders, but still require $O(L)$ steps. 
\\
\\
More recent dynamic masking strategies attempt to accelerate diffusion by prioritizing easy, or high confidence, tokens first \cite{ghazvininejad2019mask}. However, the ESM-1v paper notes that purely sampling based on confidence can lead to inconsistencies if dependent variables are unmasked simultaneously without conditioning on each other \cite{meier2021language}. EB-Sampler attempts to account for dependencies by bounding joint dependence error using marginal entropies; albeit, this relaxation collapses dependency structure into aggregate uncertainty measures, which cannot explicitly represent or correct higher-order, conditional, or asymmetric dependencies between tokens \cite{benhamu2025acceleratedsamplingmaskeddiffusion}.
\\
\\
This lack of joint structure is the gap that our work addresses: by using MI, rather than just entropy, we identify variables that are conditionally independent and can truly be sampled in parallel confidently. This work offers a theoretically grounded method for efficient parallel generation. At the same time, we demonstrate how a model's belief about the mutual information between tokens can provide interpretable prediction insight. 

\section{Preliminaries}
\subsection{Discrete Sequence Modeling}
Let $x_0 = (x_0^1, x_0^2, \cdots, x_0^L)$ be a discrete sequence of length $L$, where each token belongs to a vocabulary $V$. In MDMs, we extend this vocabulary to include a mask token: $V^+ = V \cup \{[M]\}$.

\subsection{Forward Diffusion Process}
We define a forward diffusion process that gradually corrupts the data $x_0$ into pure noise. Unlike continuous diffusion models, where Gaussian noise is added, MDMs employ an absorbing mask token state. 
\\
\\
For any timestep $t\in [0,1]$, the forward process $q(x_t|x_0)$ is defined such that each token is independently masked with probability $\gamma (t)$, where $\gamma$ is a monotomically increasing noise schedule:
\begin{equation} q(x_t^i | x_0^i) = 
\begin{cases} 
1 - \gamma(t) & \text{if } x_t^i = x_0^i \ \gamma(t) \text{ if } x_t^i = [M] \\ 
0 & \text{otherwise} 
\end{cases} 
\end{equation}
As $t\to 1$, $\gamma (t) \to 1$, and the sequence becomes fully masked. 

\subsection{Reverse Process and Training}
The generative process requires learning a reverse distribution $p_\theta(x_{t-1} | x_t)$ to denoise the sequence. In modern MDM parameterizations, as unmasked tokens are usually never resampled, the neural network $f_\theta$ is effectively trained to predict the clean data $x_0$ directly from the masked input $x_t$.
\\
\\
The model outputs a categorical distribution over the vocabulary for every position $i$:
\begin{equation}
p_\theta(x_0^i = v \mid x_t) = \text{Softmax}(f_\theta(x_t)^i)_v
\end{equation}
The network is trained to minimize the cross-entropy loss between the predicted distribution and the ground truth tokens at masked positions:
\begin{equation}
\mathcal{L}_\text{MDM} = \mathbb{E}_{t \sim \mathcal{U}(0,1), x_0 \sim \mathcal{D}} \left[ \sum_{i: x_t^i = [M]} -\log p_\theta(x_0^i \mid x_t) \right]
\end{equation}

\begin{figure}
    \centering
    \includegraphics[width=1.0\linewidth]{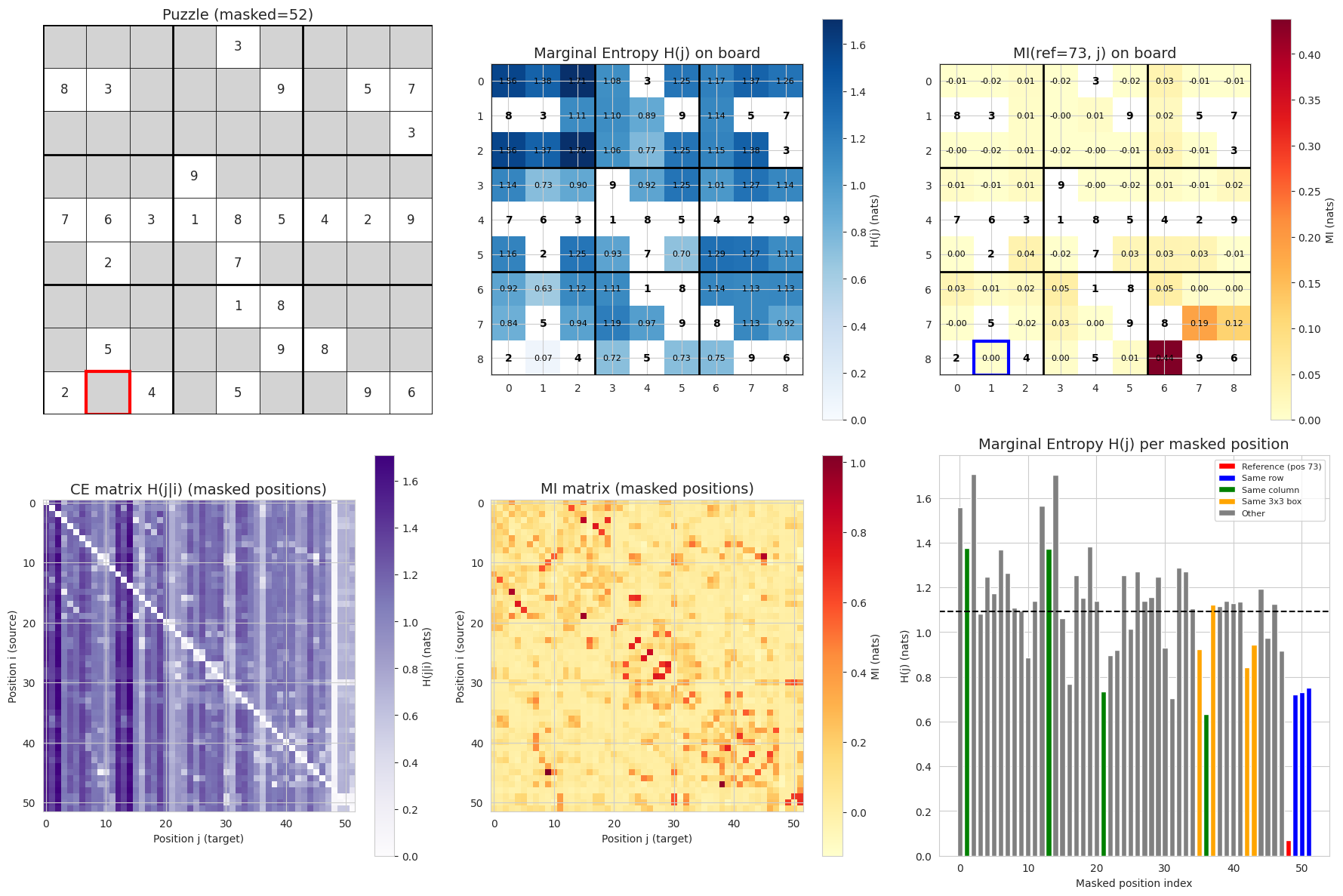}
    \caption{
    Visualization of the MDM's beliefs before unmasking an 8 in the boxed square.\\
    \textbf{Top Right: } `1' must lie in two cells of the bottom row, resulting in high MI (0.44 nats)
    }
    \label{fig:sudoku_interp}
\end{figure}

\subsection{From Marginals to Pairwise Dependencies}

Standard MDM training effectively learns the marginal conditional distributions $p(x^i \mid x_t)$. While these probabilities are sufficient for order agnostic sampling, they do not explicitly quantify the dependency structure between unmasked variables. Therefore, efficient parallel sampling requires identifying subsets of variables that are conditionally independent given the context. To achieve this goal, we must move beyond the marginal predictions. \\

The mutual information between positions $x_i$ and $x_j$ is the KL divergence between the joint distribution $p(x_i, x_j)$ and the product of the marginal distributions $p(x_i)p(x_j)$. This info-theoretic quantity helps us quantify dependence between the model's beliefs about two positions.

As shown in Fig.~\ref{fig:sudoku_interp}, although marginal entropies vary across the board, they do not reveal which cells can be resolved independently. In contrast, the MI map in exposes strong pairwise dependencies aligned with Sudoku constraints, demonstrating that the model’s internal belief captures relational structure beyond per-cell uncertainty.

\section{Methodology}
Our approach hinges on the ability to quantify pairwise dependencies between masked variables efficiently. We first define an expensive but exact method to calculate ground truth MI from the pretrained MDM. This ground truth MI is the model's belief of the MI between different tokens in the sequence. We then train a lightweight neural estimator to approximate this quantity directly from hidden states, enabling efficient MI-guided parallel sampling. Our approach can be viewed as adding a computation head to one of the final layers of the pretrained MDM. 

\subsection{Ground Truth Mutual Information Calculation}
To train a predictor, we first require ground truth MI values. For a discrete sequence model with context $C$, the set of currently unmasked tokens, the pairwise mutual information between two masked positions $i$ and $j$ is defined as:
\begin{equation}
I(X_i; X_j \mid C) = \sum_{x_i} \sum_{x_j} P(x_i, x_j \mid C) \log \frac{P(x_i, x_j \mid C)}{P(x_i \mid C)P(x_j \mid C)}
\end{equation}
Computing this exactly requires the joint distribution $P(X_i, X_j \mid C)$. Since MDMs typically output marginals $P(X_i \mid C)$, we employ a perturbation-based strategy to recover the joint.
\\
\\
We compute the ground truth via a brute-force conditional probing method, requiring $1 + N \cdot |V|$ forward passes, where $N$ is the sequence length and $|V|$ is the vocabulary size:
\begin{enumerate}
    \item \textbf{Base Pass:} Run the model on the masked sequence $C$ to obtain marginals $P(X_i \mid C)$ for all $i$. From this, we compute individual entropies $H(X_i \mid C)$.
    \item \textbf{Conditional Passes:} For each position $i$ and every possible token $v \in V$, we temporarily fix $X_i = v$ and run a forward pass. This yields the conditional distribution $P(X_j \mid X_i=v, C)$ for all $j \neq i$.
    \item \textbf{Conditional Entropy:} We can compute the conditional entropy $H(X_j \mid X_i, C)$ by summing the pointwise conditional entropies with  $X_i = v$.\item \textbf{MI Calculation:} We compute the MI as the reduction in entropy:$$ I(X_i; X_j \mid C) = H(X_j \mid C) - H(X_j \mid X_i, C) $$
\end{enumerate}
This process generates a ground truth matrix $M_{GT} \in \mathbb{R}^{N \times N}$ for a given context.

\subsection{Neural MI Estimator}
Since the $O(N \cdot |V|)$ cost of ground truth calculation is very expensive during inference, we train a neural estimator $f_\phi$ to predict the pairwise MI matrix directly from the MDM's hidden representations.

\paragraph{Architecture and Input}
The estimator operates on the hidden states $h \in \mathbb{R}^{N \times D}$ produced by the frozen MDM backbone. The model outputs a symmetric matrix $\hat{I} \in \mathbb{R}^{N \times N}$ representing the estimated pairwise MI for all positions simultaneously.
$$ \hat{I} = f_\phi(\text{MDM}(X_t)) $$

\paragraph{Training Objective}
We generate training data by sampling random noise levels $t \sim \mathcal{U}[0, 1]$. For a sequence $X_0$, we generate a masked input $X_t = \text{sample\_noise}(X_0, t)$. We compute the ground truth matrix $M_{GT}$ and train the estimator to minimize the Mean Squared Error (MSE) over the masked indices:
\begin{equation}
\mathcal{L}_{MI} = || M_{GT} - \hat{I} ||_F^2
\end{equation}

\subsection{MI-Guided Parallel Sampling}
Standard parallel decoding strategies such as Mask-Predict unmask the top-$k$ tokens based solely on confidence, or lowest entropy. However, if high-confidence tokens are highly correlated (i.e. they have high MI), unmasking them simultaneously without conditioning on each other leads to inconsistencies (e.g. placing two 3's in the same Sudoku row). We propose an MI-guided greedy selection strategy. The goal is to select a batch of tokens $S$ to unmask such that they are mutually independent. \\
\begin{algorithm}[h]
\caption{Budgeted MI-Guided Sampling}
\begin{algorithmic}[1]
\REQUIRE Masked indices $\mathcal{M}$, predicted MI matrix $\hat{I}$, Budget $\gamma$, Penalty $\lambda$
\STATE Compute per-token entropies $h_i = H(p_\theta(x_i \mid C))$ for all $i \in \mathcal{M}$
\STATE Sort $\mathcal{M}$ by increasing entropy $h_i$ (highest confidence first)
\STATE Initialize selected set $U \leftarrow \emptyset$
\STATE Initialize remaining budget $B \leftarrow \gamma$

\FOR{candidate $i$ in sorted $\mathcal{M}$}
    \STATE \textbf{Estimate Dependency Cost:}
    \STATE $d(i \mid U) \leftarrow \sum_{j \in U} \hat{I}_{i,j}$
    
    \STATE \textbf{Calculate Total Cost:}
    \STATE $cost \leftarrow h_i + \lambda \cdot d(i \mid U)$
    
    \IF{$cost \leq B$}
        \STATE $U \leftarrow U \cup \{i\}$
        \STATE $B \leftarrow B - cost$
    \ENDIF
    
    \STATE \textit{// Optional: Break if budget is too low to accept any future candidate}
    \IF{$B \le 0$} \STATE \textbf{break} \ENDIF
\ENDFOR

\STATE \textbf{Unmask:} Sample tokens in $U$ based on marginals
\STATE \textbf{return} Updated sequence
\end{algorithmic}
\end{algorithm}

This strategy ensures that we only parallelize steps where the model believes the variables are conditionally independent. If $\hat{I}_{i,j}$ is high, the algorithm forces sequential processing: once $i$ is sampled, the distribution for $j$ will update in the next step, preserving consistency.






\section{Experiments}

\subsection{Experiment Setup and Estimator Training}
\begin{itemize}
    \item \textbf{Sudoku:} We trained a masked diffusion model with 4,158,346 parameters on a dataset of 100,000 randomly generated sudoku puzzles. We also trained a mutual information predictor head of 99,969 parameters, over 10 epochs.
    \item \textbf{Proteins:} We use ESM-Cambrian with 300M parameters, a protein language model developed by Evolutionary Scale Modeling. We then trained a small predictor head of $\sim$810K parameters, over 10000 proteins and 5 epochs.
\end{itemize}

\subsection{Sudoku: Parallelization Experiment}

We evaluated MI-guided parallel sampling of cells versus naive sequential/parallel sampling and EB-Sampler on a set of 1000 unseen hard Sudoku puzzles.

\begin{table}[t]
\caption{Parallelization performance for Sudoku generation. Our MI-Guided Sampler achieves higher accuracy than the sequential baseline while requiring significantly fewer forward passes.}
\label{tab:sudoku_parallel}
\begin{center}
\begin{small}
\begin{sc}
\begin{tabular*}{\columnwidth}{@{\extracolsep{\fill}} lcc}
\toprule
Method & Avg. Passes & Sol. Acc. \\
\midrule
Sequential & 53.9 & 61.6\% \\
Naive ($k=4$) & 14.9 & 52.4\% \\
Naive ($k=7$) & 9.0 & 36.8\% \\
\midrule
EB-Sampler ($\gamma=0.2$) & 15.3 & 61.0\% \\
EB-Sampler ($\gamma=0.5$) & 9.9 & 51.2\% \\
\midrule
\textbf{MI-Guided ($\gamma=0.3$)} & \textbf{15.2} & \textbf{63.6\%} \\
MI-Guided ($\gamma=0.6$) & 9.7 & 56.2\% \\
\bottomrule
\end{tabular*}
\end{sc}
\end{small}
\end{center}
\vskip -0.1in
\end{table}

\begin{figure}[H]
    \centering
    \includegraphics[width=0.8\linewidth]{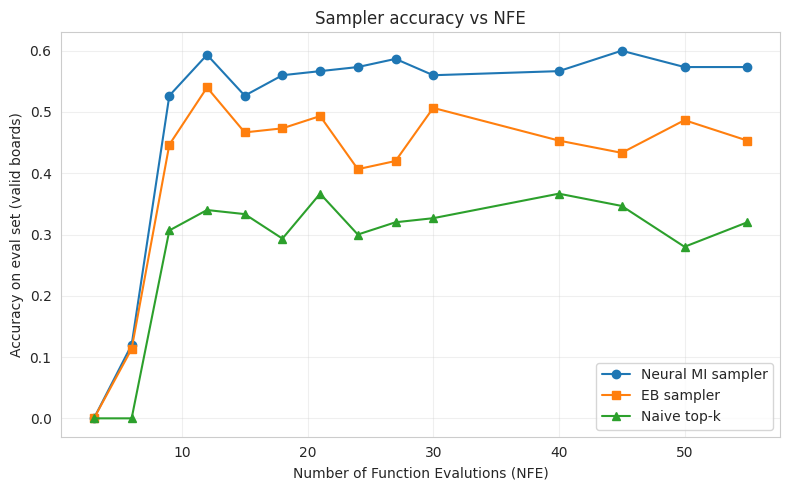}
    \caption{Sudoku, accuracy vs NFE}
    \label{fig:placeholder}
\end{figure}

\subsection{Protein Sequences: MI-guided Parallelization}

Using a variety of techniques, we unconditionally generated 500 random proteins of lengths 50-100. We also embedded 500 reference samples from the UniRef50 database, of length 50-100. We then clustered them with k-means (fit on the reference, k=15), and computed their Jensen-Shannon divergence to the reference set.

\begin{table}[t] 
\caption{Comparison of sequence generation methods for ESM-C. MI-guided sampling achieves a better speed-accuracy trade-off compared to naive parallel baselines.}
\label{tab:protein_parallel}
\begin{center}
\begin{small}
\begin{sc}
\begin{tabular*}{\columnwidth}{@{\extracolsep{\fill}} lcc}
\toprule
Method & Avg. Passes & JSD to Ref. \\
\midrule
Sequential & 74.8 & 0.093 \\
Naive ($k=4$) & 19.1 & 0.185 \\
Naive ($k=8$) & 9.8 & 0.196 \\
Naive ($k=12$) & 6.2 & 0.218 \\
\midrule
\textbf{MI-guided ($\gamma=2, \lambda=1$)} & \textbf{15.3} & \textbf{0.136} \\
MI-guided ($\gamma=4, \lambda=1$) & 10.0 & 0.174 \\
\bottomrule
\end{tabular*}
\end{sc}
\end{small}
\end{center}
\vskip -0.1in 
\end{table}

\begin{figure}[H]
    \centering
    \includegraphics[width=0.8\linewidth]{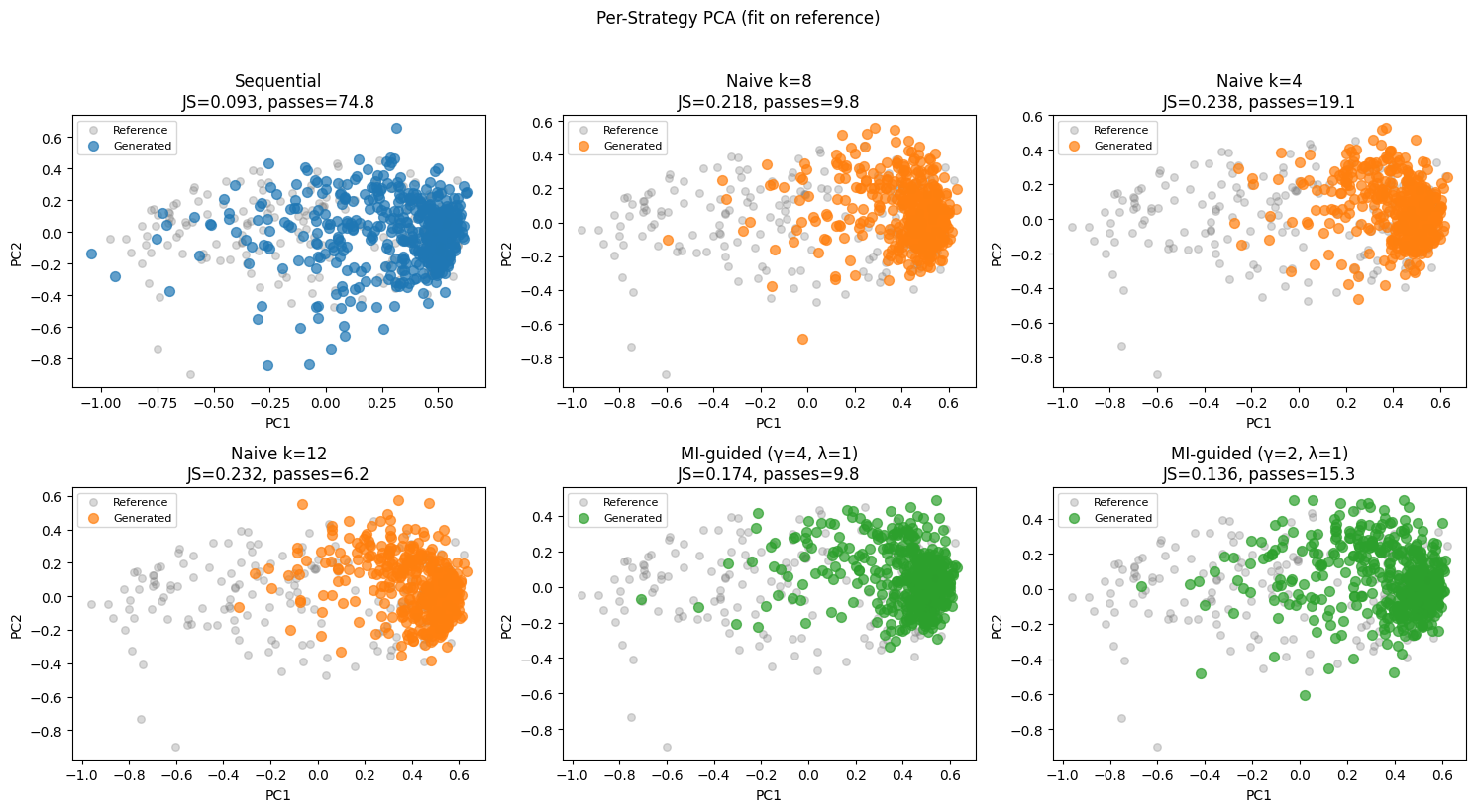}
    \caption{PCA projections of ESM-C embeddings of various sampling methods.}
    \label{fig:placeholder}
\end{figure}

\section{Discussion}

Mutual information guided parallelization significantly reduces average forward passes in generation in masked diffusion models. Although this speedup generally comes with a degradation in the quality of generated samples, our experiments show a substantial improvement beyond naive parallelization, achieving efficiency gains without a substantial degradation in quality.

However, there are limitations with our approach --- our predictor is far from perfect, and requires substantial amount of preparation to setup and train. Future work can be done to investigate optimal architectures for the predictor head, as well as improved curriculum / training strategies beyond computing ground truth mutual information on the fly for each example.

\section{Conclusion}
In this work, we presented a novel framework for estimating pairwise MI within MDMs, bypassing the computationally expensive ground truth calculation required by traditional methods. Our results demonstrate that this neural estimator serves as a powerful dual-purpose tool for both model interpretability and efficient generation. In the domain of structured logic (Sudoku), we showed that our estimator effectively uncovers the model's learned internal representations, revealing that MDMs naturally acquire the rigid constraints of the game (row, column, and box rules) without explicit programming. Furthermore, our application to protein sequence generation highlights the practical utility of MI-guided sampling. By dynamically identifying conditionally independent tokens, our method achieves an ideal balance between speed and performance. Specifically, we demonstrated that MI-guided sampling can reduce the number of forward passes by nearly an order of magnitude compared to sequential methods, while significantly outperforming naive parallel baselines in generative quality. Ultimately, this work suggests that explicitly modeling variable dependencies is key to unlocking the full potential of discrete diffusion models. It bridges the gap between the high quality of sequential sampling and the efficiency of parallel decoding, offering a theoretically grounded path toward faster, more reliable discrete generation.

\section*{Impact Statement}
This paper presents work whose goal is to advance the field of Machine Learning. There are many potential societal consequences of our work, none which we feel must be specifically highlighted here.

\section*{Acknowledgements}
We would like to thank Prof. Kannan Ramchandran and Justin Kang for their guidance during the graduate EE 229A Information Theory course at UC Berkeley.

\bibliography{refs}
\bibliographystyle{icml2026}

\end{document}